\documentclass[11pt]{article}

\usepackage[preprint]{acl}

\usepackage{times}
\usepackage{latexsym}

\usepackage[T1]{fontenc}

\usepackage[utf8]{inputenc}

\usepackage{microtype}

\usepackage{inconsolata}

\usepackage{graphicx}

\usepackage{amsmath}
\usepackage{subcaption}
\usepackage{booktabs}
\usepackage{multirow}

\usepackage{amsmath}
\usepackage{amssymb}

\newcommand{\think}[1]{\textcolor{blue}{\texttt{<think>}} #1 \textcolor{blue}{\texttt{</think>}}}

\newcommand{\answer}[1]{\textcolor{purple}{\textbf{\texttt{<answer>}}} {\textbf{#1}} \textcolor{purple}{\textbf{\texttt{</answer>}}}}

\usepackage{xspace}
\newcommand{\method}{\texttt{R$^2$PO}\xspace}

\usepackage{stfloats}

\title{R$^2$PO: Decoupling Rollout and Inference Policies for LLM Reasoning}

\author{
    Jingchu Wang$^{1,2}$,
    Bingbing Xu$^{1}$\thanks{\; Corresponding author.}, \\ 
    \textbf{Yige Yuan}$^{1,2}$, 
    \textbf{Dan Zhang}$^{3}$,
    \textbf{Bin Xie}$^{1,2}$,
    \textbf{Xiaoqian Sun}$^{1}$,
    \textbf{Huawei Shen}$^{1}$\\
        $^{1}$State Key Laboratory of AI Safety, Institute of Computing Technology, CAS, \\
        $^{2}$University of Chinese Academy of Sciences \\
        $^{3}$National University of Singapore \\
        \texttt{\{wangjingchu25e,xubingbing\}@ict.ac.cn}
}

\begin{document}
\maketitle
\begin{abstract}
Existing reinforcement learning methods for LLM reasoning implicitly assume that the policy generating training trajectories should coincide with the one producing inference responses. We argue that this is a misleading inductive bias: the \emph{optimization-optimal} trajectory distribution favors informative gradients, whereas the \emph{inference-optimal} response distribution emphasizes accuracy and consistency. Forcing both into a single policy entangles their gradients and suppresses exploration. We propose \textbf{\method} (\textbf{R}esidual \textbf{R}ollout \textbf{P}olicy \textbf{O}ptimization), which attaches a lightweight Residual Rollout-Head atop the policy to decouple training trajectories from inference responses, diversifying rollouts during training while keeping inference generation intact. Experiments show that \method consistently outperforms baselines, with average accuracy gains of \textbf{3.4\%} on MATH-500 and \textbf{1.3\%} on APPS, alongside more diverse rollouts and reduced length bias. Our code is available at \href{https://github.com/RRPO-ARR/Code}{\color[rgb]{0.857, 0.251, 0.296}{this link}}.
\end{abstract}

\section{Introduction}

\begin{figure*}
    \centering
    \includegraphics[width=0.9\linewidth]{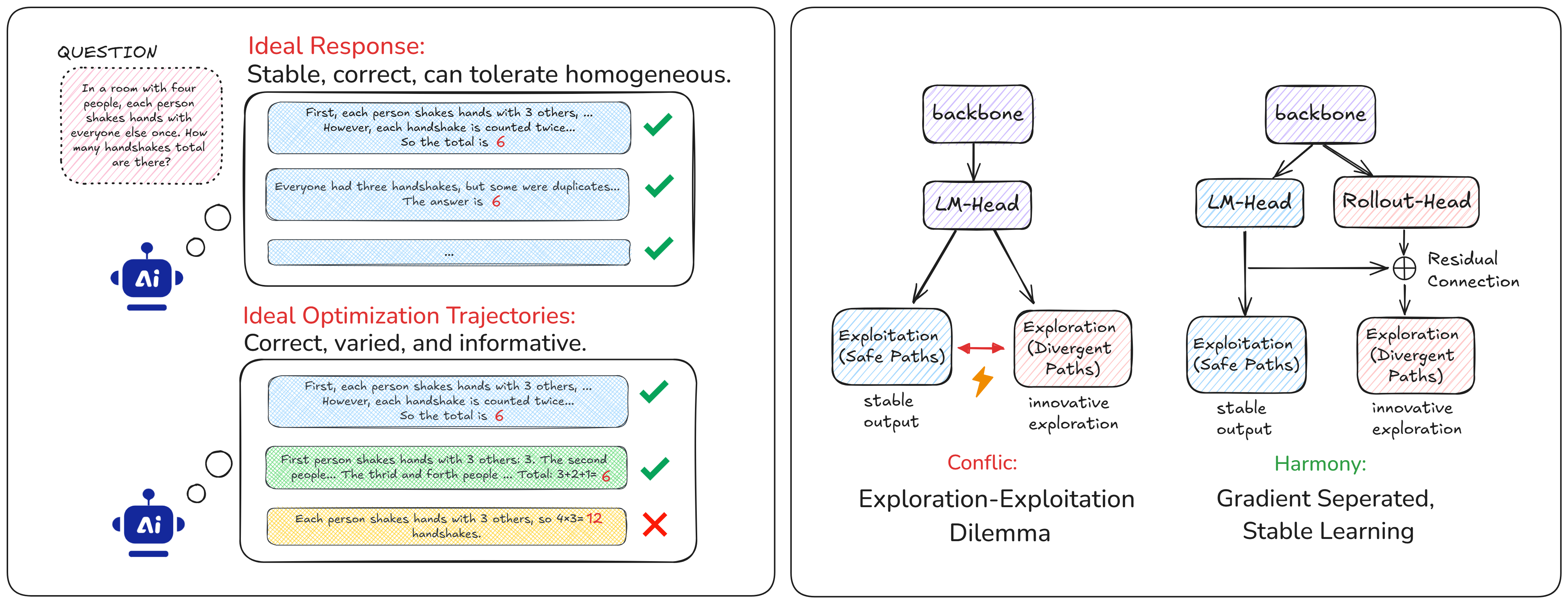}
    \caption{\textbf{Decoupling Ideal Responses and Optimization Trajectories.} Left: The mismatch between ideal inference responses and ideal training trajectories. Right: Single-head policy optimization versus \method with a decoupled Rollout-Head for stable exploration.}
    \label{fig:intro}
\end{figure*}

In recent years, Reinforcement Learning (RL) for Large Language Models (LLMs) has become a central paradigm for improving their reasoning capabilities. It underpins approaches such as Reinforcement Learning from Human Feedback (RLHF)~\cite{DBLP:conf/nips/ChristianoLBMLA17, ouyang2022training} and Verifiable-Reward-based fine-tuning (RLVR)~\cite{DBLP:journals/corr/abs-2411-15124, DBLP:journals/corr/abs-2501-12948}, and has achieved remarkable success in complex reasoning tasks such as mathematical problem solving ang code generation~\cite{deepseekai2025deepseekv3technicalreport, DBLP:journals/corr/abs-2503-09501, ouyang2022training, DBLP:journals/corr/abs-2501-12948, DBLP:journals/corr/abs-2503-14476}. By optimizing models directly against task-specific or preference-based reward signals, RL enables LLMs to move beyond pure imitation and exhibit stronger performance on complex reasoning tasks~\cite{DBLP:conf/nips/StiennonO0ZLVRA20, DBLP:conf/nips/ShinnCGNY23}.


A common but rarely questioned assumption underlies these methods (e.g., PPO~\cite{DBLP:journals/corr/SchulmanWDRK17}, GRPO~\cite{DBLP:journals/corr/abs-2402-03300}): the policy used to generate \emph{training rollouts} is identical to the policy used to produce \emph{inference responses}. We argue that this is a misleading inductive bias for reasoning RL. As illustrated in Figure~\ref{fig:intro} (left), the \emph{inference-optimal} distribution emphasizes accuracy and consistency and thus tolerates homogeneous outputs, whereas the \emph{optimization-optimal} distribution requires diversity and broad solution-space coverage, including informative erroneous paths that provide effective gradient signals. Tying these two roles to a single set of parameters forces the optimization process to be dominated by response-level correctness, which suppresses effective exploration for diversity. As a result, the model frequently collapses into self-repetition during training~\cite{DBLP:conf/nips/MoallaMPPG24, DBLP:journals/corr/abs-2505-22617}.

To overcome this, we propose \textbf{\method} (Residual Rollout Policy Optimization). Unlike paradigms that compress the two conflicting objectives into a single policy, we introduce a lightweight residual module atop the frozen backbone to decouple inference responses and training trajectories. 
The Residual Rollout-Head learns to make controlled adjustments to the base distribution, exploring candidate trajectories that the main policy would otherwise downweight.
This decoupling keeps the primary policy stable at inference time, while ensuring diverse trajectory exploration during training via the residual Rollout-Head.
The Rollout-Head can be discarded at inference, ensuring efficient and consistent deployment. 

We investigate the effectiveness of \method across mathematical reasoning and code generation tasks on Qwen2.5-3B and Qwen3-8B. Experimental results underscore that \method outperforms the standard GRPO baseline, achieving an average accuracy increment of \textbf{3.4\%} on MATH-500 and \textbf{1.3\%} on APPS. We further reveal that our framework produces more diverse training rollouts and mitigates length bias, maintaining a stable reasoning trajectory even in the late stages of optimization.


The contributions of this work are as follows: 
\begin{itemize} 
    \item \textbf{A new perspective on reasoning RL.} We identify and challenge a misleading inductive bias in existing methods: the rollout policy need not coincide with the inference policy. We articulate why the \emph{optimization-optimal} and \emph{inference-optimal} trajectory distributions are intrinsically different.
    \item \textbf{A lightweight framework, \method.} We realize this principle with a lightweight residual Rollout-Head atop a frozen backbone, jointly trained but discarded at inference, fully compatible with standard GRPO.
    \item \textbf{Consistent empirical gains.} Across mathematical reasoning and code generation benchmarks, \method improves accuracy, increases rollout diversity, controls length bias, and stabilizes training, with negligible overhead.
\end{itemize}

\section{Related Works}
We review related work from tow perspectives: 
(1) RL for reasoning-oriented LLMs, and
(2) exploration mechanisms in LLM RL.

\paragraph{RL for Reasoning LLMs} LLMs have demonstrated significant reasoning improvements through Reinforcement Learning from Human Feedback (RLHF) \cite{ouyang2022training}. While traditional PPO-based frameworks \cite{DBLP:journals/corr/SchulmanWDRK17} rely on separate value models, recent paradigms like Reinforcement Learning with Verifiable Rewards (RLVR) \cite{lambert2025tulu3pushingfrontiers, DBLP:journals/corr/abs-2501-12948} have gained prominence in domains like mathematics and coding. However, these verifiable reward landscapes are often sparse and unforgiving, posing significant exploration and optimization challenges for reasoning RL. 

\paragraph{Exploration Mechanism in LLM RL}
\label{sec:ee}

Balancing exploration and exploitation is a fundamental challenge in reinforcement learning~\cite{sutton2018reinforcement}. Traditional approaches such as Soft Actor-Critic (SAC)~\cite{DBLP:conf/icml/HaarnojaZAL18} employ entropy regularization to sustain exploration and prevent premature convergence, and in LLM fine-tuning most paradigms~\cite{DBLP:conf/nips/StiennonO0ZLVRA20, deepseekai2025deepseekv3technicalreport} rely on KL-divergence penalties to implicitly preserve diversity. However, such soft regularization is often insufficient, as policy entropy can still collapse rapidly, leading to suboptimal solutions~\cite{DBLP:conf/nips/MoallaMPPG24, DBLP:journals/corr/abs-2505-22617}. Several recent studies further investigate the role of entropy in LLM RL: \cite{DBLP:journals/corr/abs-2509-26114} analyzes the interaction between PPO clipping and entropy dynamics; \cite{DBLP:journals/corr/abs-2506-14758} examines how entropy-characterized exploration affects the depth and diversity of reasoning trajectories; and other works incorporate entropy-related signals into the objective itself, such as entropy-regularized policy optimization~\cite{DBLP:journals/corr/abs-2509-22576} and entropy-aware token-level updates~\cite{wen2024entropyregularizedtokenlevelpolicyoptimization}. All of these methods, however, modify the objective while still using a single policy for both rollout and inference, which indeed mirrors the gap between the ideal optimization trajectory and the ideal inference response that motivates our work.

\section{Method}

Motivated by the gap between the ideal inference response and the ideal training trajectory, we propose \method, a framework that decouples these two roles instead of forcing them into a single shared policy. We first analyze why a single policy cannot reconcile both objectives, and then introduce the overall framework, policy implementation, reward design, and iterative training procedure of \method.

\subsection{Motivation}
\label{sec:method:motivation}

In standard RL fine-tuning for LLM reasoning, a single policy $\pi_\theta$ actually plays two roles at the same time. During training, $\pi_\theta$ generates rollouts whose rewards drive parameter updates; and during inference, the same $\pi_\theta$ produces responses that are scored as the system's output. 

These two roles (illustrated in Figure~\ref{fig:intro}), however, are governed by fundamentally different objectives. The inference role aims to maximize the expected task reward of the deployed response,
\begin{equation}
\small
    \mathcal{O}_{\mathrm{infer}}(\pi)
    = \mathbb{E}_{x \sim \mathcal{D},\, y \sim \pi(\cdot \mid x)}\bigl[\,r(x, y)\,\bigr],
\end{equation}
which favors low-entropy, mode-seeking behavior. The training role aims to maximize the magnitude of the gradient signal produced by rollouts, defined as $\mathcal{O}_{\mathrm{train}}$, which under group-relative advantages requires rollout distributions that spread across trajectories with different rewards so that the per-group advantage
\begin{equation}
\small
    \hat{A}_i
    = \frac{r(x, y_i) - \mathrm{mean}\bigl(\{r(x, y_j)\}_{j=1}^{G}\bigr)}{\mathrm{std}\bigl(\{r(x, y_j)\}_{j=1}^{G}\bigr)}
\end{equation}
does not collapse to zero. Consequently, training often benefits from rollout distributions that remain diverse, exploratory, and tolerant of occasionally incorrect trajectories.

When the same parameters carry both objectives, gradients induced by response-level correctness progressively suppress deviations that could otherwise reveal alternative reasoning paths. As a result, the policy tends to reinforce a narrow subset of safe trajectories, limiting the discovery of rare but informative behaviors.

These observations suggest that the two roles should not be carried by a fully shared parameter set. Rather than insisting on one policy that simultaneously satisfies both objectives, we allocate a small set of parameters to absorb $\mathcal{O}_{\mathrm{train}}$, while keeping the deployed response policy dedicated to inference. This principle of partial parameter decoupling is what \method instantiates.

\begin{figure}
    \centering
    \includegraphics[width=1\linewidth]{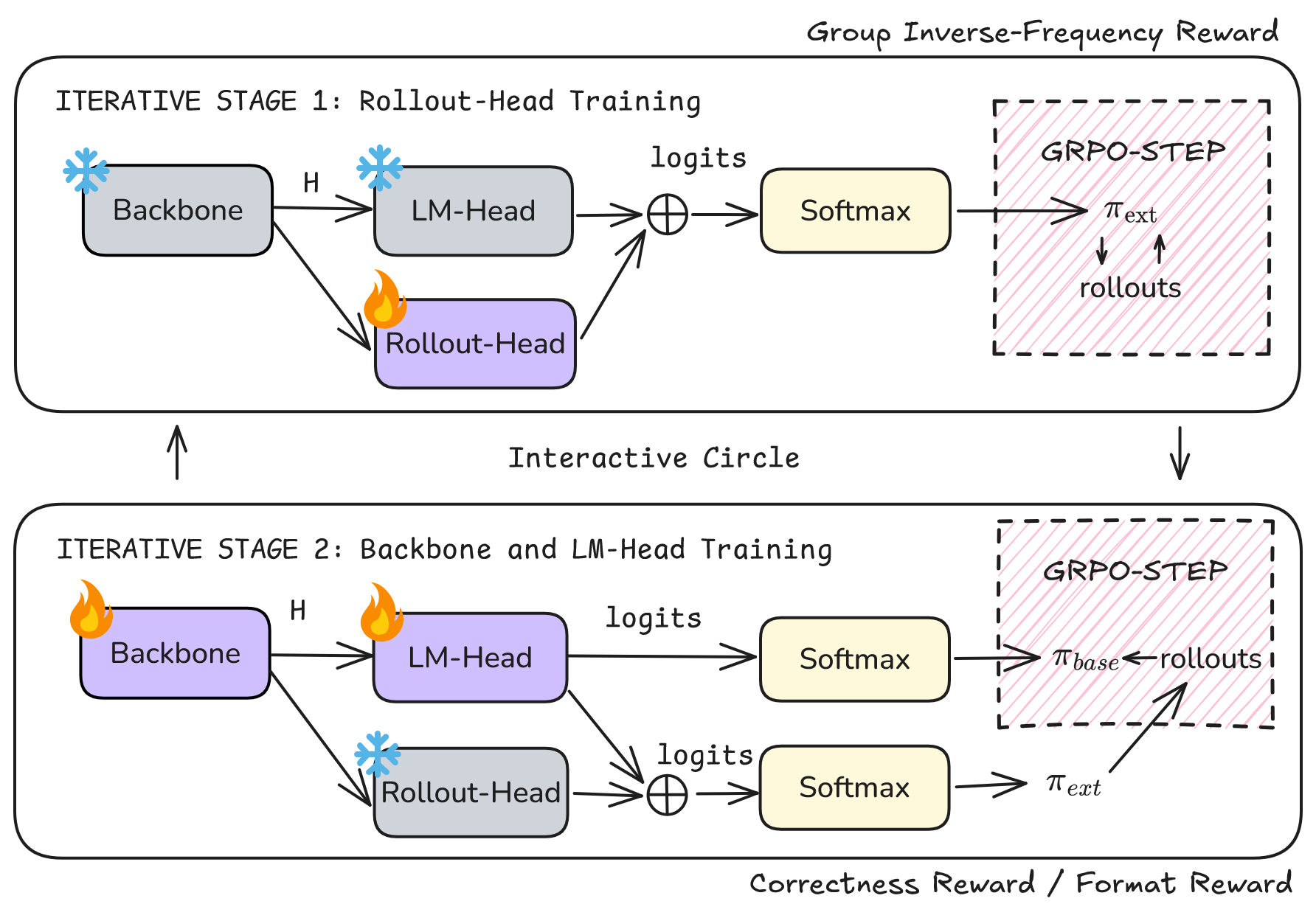}
\caption{Overview of \textbf{\method}. An iterative loop of two alternating stages: Stage 1 (top) optimizes the Rollout-Head using GIF reward with a frozen backbone; Stage 2 (bottom) updates the base policy on task rewards using the Rollout-Head as a fixed behavioral sampler.}
    \label{fig:iterative_train}
\end{figure}

\subsection{Overall Framework}
\label{sec:method:framework}

To decouple the two optimization objectives identified above, we propose \method, whose overall framework is clearly illustrated in Figure~\ref{fig:iterative_train}. \method attaches a lightweight residual head, called the \textbf{Rollout-Head}, that is optimized only against the training objective $\mathcal{O}_{\mathrm{train}}$, while original backbone and language modeling head (LM head) remain responsible for the inference objective $\mathcal{O}_{\mathrm{infer}}$. This yields two policies sharing the same backbone: main policy $\pi_\theta$, produced by the original LM Head, which serves as the deployed inference policy; and rollout policy $\pi_\phi$, produced by combining the LM Head with the Rollout-Head, which serves as the behavioral sampler during training. To coordinate the two policies, \method adopts an iterative training scheme that alternates between optimizing the Rollout-Head for exploration and updating the main policy from the resulting rollouts.

\subsection{Dual-Policy Implementation}
\label{sec:method:impl}

We implement the Rollout-Head as a residual branch attached on top of the transformer backbone. Concretely, it is a two-layer Multi-Layer Perceptron (MLP) that takes the backbone hidden states $H \in \mathbb{R}^{d}$ as input and produces a logit-level offset over the vocabulary space $\mathbb{R}^{V}$, where $d$ is the hidden dimension and $V$ is the vocabulary size. Let $f_{\mathrm{LM}}$ denote the original LM Head and $f_{\mathrm{RO}}$ the Rollout-Head. The two policies share the backbone and differ only in their output heads:
\begin{equation}
\begin{aligned}
    \pi_{\theta} &= \mathrm{Softmax}\bigl(f_{\mathrm{LM}}(H)\bigr), \\
    \pi_{\phi}   &= \mathrm{Softmax}\bigl(f_{\mathrm{RO}}(H) + f_{\mathrm{LM}}(H)\bigr).
\end{aligned}
\end{equation}
This concrete formulation operationalizes the two-stage scheme described in Section~\ref{sec:method:framework}: during Stage 1, we optimize the residual parameters $\phi$ of $f_{\mathrm{RO}}$ while anchoring it around the frozen base logits; during Stage 2, $f_{\mathrm{RO}}$ is fixed to output stable exploratory perturbations that guide the optimization of the backbone and $\theta$. 

    In our framework, the main policy $\pi_\theta$ is responsible for stable exploitation and is the only policy used at inference time, while the rollout policy $\pi_\phi$ incorporates the residual perturbation introduced by the Rollout-Head and is used exclusively during training. To stabilize the cold-start phase, we zero-initialize the Rollout-Head, so that $\pi_\phi$ coincides with $\pi_\theta$ at the beginning of training and gradually deviates from it as exploration is learned. Because the perturbation is restricted to a separate residual branch, the Rollout-Head can be discarded after training, leaving the deployed model architecturally identical to a standard LLM.

\subsection{Reward Design}
\label{sec:method:reward}

The two policies are trained with different rewards, reflecting their different objectives.

\paragraph{Group Inverse-Frequency Reward.}
Because the Rollout-Head is responsible for the training-time objective $\mathcal{O}_{\mathrm{train}}$, its reward should encourage rollouts that yield non-trivial group-relative advantages: a group whose rollouts collapse into a single reward value provides no learning signal under group-relative advantage estimation. We therefore upweight trajectories whose reward outcome is rare within the group, reflecting the intuition of intrinsic-motivation and novelty-search literature~\cite{DBLP:conf/nips/BellemareSOSSM16, DBLP:journals/ec/LehmanS11}. Let $\mathcal{G} = \{y_1, \dots, y_G\}$ be a group of rollouts sharing the same query with verifiable rewards $r_i$, and let $\mathcal{B}_k = \{y_i \in \mathcal{G} \mid r_i = v_k\}$ denote the bin of rollouts sharing the unique reward value $v_k$. The Group Inverse-Frequency (GIF) score for $y_i \in \mathcal{B}_k$ is the inverse of the bin's cardinality:
\begin{equation}
    s_i = \frac{1}{|\mathcal{B}_k|} = \frac{1}{\sum_{j=1}^{G} \mathbf{1}[r_j = r_i]}.
\end{equation}
To prevent gradient explosion from groups with extreme bin imbalance, we apply Z-score standardization within each group to obtain the final exploration reward $R_{\mathrm{GIF}}$.

\paragraph{Correctness and Formatting Reward.}
Because the main policy $\pi_\theta$ is responsible for the inference-time objective $\mathcal{O}_{\mathrm{infer}}$, its reward must reflect the actual quality of the deployed response on downstream task. Following recent advances in reasoning model training~\cite{DBLP:conf/nips/Le0GSH22, DBLP:conf/nips/ShinnCGNY23, DBLP:conf/icml/GehringZCMCS25}, we adopt a rule-based reward that provides dense and verifiable feedback without a learned reward model. For a query $x$ and generated response $y_i$, the total task reward is
\begin{equation}
\small
    r(x, y_i) = \mathbf{1}_{\mathrm{correct}}(y_i) \cdot R_{\mathrm{acc}} + \mathbf{1}_{\mathrm{format}}(y_i) \cdot R_{\mathrm{fmt}},
\end{equation}
where $\mathbf{1}_{\mathrm{correct}}$ and $\mathbf{1}_{\mathrm{format}}$ are indicator functions checking numerical correctness and adherence to the formatting template (e.g., enclosing the final answer between \texttt{<answer>} and \texttt{</answer>}). This deterministic reward keeps the optimization signal of the main policy consistent during training.

\subsection{Iterative Training}
\label{sec:method:training}

\method adopts a two-stage alternating optimization scheme to separate the update signals for exploration and exploitation. Both stages utilize the standard Group Relative Policy Optimization (GRPO) \cite{DBLP:journals/corr/abs-2402-03300} objective:

\begin{small}
\begin{equation}
\begin{aligned}
\mathcal{J}_{\mathrm{GRPO}}(\theta) &= \mathbb{E}_{x \sim \mathcal{D}, \{y_i\}_{i=1}^G \sim \pi_{\theta_{\mathrm{old}}}} \left[ \frac{1}{G} \sum_{i=1}^{G} \frac{1}{|y_i|} \sum_{t=1}^{|y_i|} \right. \\
& \min \left( r_{i,t}(\theta) \hat{A}_{i,t}, \mathrm{clip}(r_{i,t}(\theta), 1-\epsilon, 1+\epsilon) \hat{A}_{i,t} \right) \\
& \left. \vphantom{\sum_{i=1}^G} - \beta \mathbb{D}_{KL}(\pi_\theta \| \pi_{ref}) \right],
\end{aligned}
\end{equation}
\end{small}

where $G$ is the number of generated responses per query $x$. The importance ratio $r_{i,t}(\theta)$ and the advantage $\hat{A}_{i,t}$ of token $y_{i,t}$ are given by:
\[
r_{i,t}(\theta) = \frac{\pi_\theta(y_{i,t} \mid x, y_{i,<t})}{\pi_{\theta_{\mathrm{old}}}(y_{i,t} \mid x, y_{i,<t})}
\]
\[
\hat{A}_{i,t} = \hat{A}_i = \frac{r(x, y_i) - \mathrm{mean}(\{ r(x,y_i)\}_{i=1}^G)}{\mathrm{std}(\{ r(x,y_i)\}_{i=1}^G)},
\]
respectively, where all the tokens in $y_i$ share the same advantage as $\hat{A}_i$. $\pi_{\mathrm{ref}}$ typically refers to the initial SFT model used for KL regularization, while $\pi_{\mathrm{old}}$ denotes the policy from the previous iteration used for importance sampling.

\subsubsection{Rollout-Head Optimization}
In Rollout-Head optimization stage, we only optimize the parameters of the Rollout-Head ($\phi$) while keeping the backbone and the primary LM Head frozen. Here, $\pi_{\phi}$ serves as both the trajectory generator and the optimization target. Let $y$ denote the trajectories sampled from $\pi_{\phi}$, the optimization objective for Stage 1 is formulated as:  
\begin{equation}
    \max \mathcal{J}_{\mathrm{GRPO}}(\phi), \quad y \sim \pi_{\phi} 
\end{equation}

In our primary experiments, we employ the Group Inverse-Frequency Reward as the default exploration signal, as it explicitly encourages diversity in optimization trajectories. A detailed comparative analysis between this reward and alternative designs is provided in Section~\ref{sec:analysis}.
The Rollout-Head is reward-agnostic; GIF is one effective instantiation, but any task-appropriate exploration signal can be substituted.

\subsubsection{Main Policy Optimization}

In the main policy optimization stage, we freeze the Rollout-Head and update the backbone and LM Head ($\theta$). Crucially, the frozen Rollout-Head acts exclusively as a \textbf{behavioral sampler} to provide diverse trajectories that the base policy might not otherwise discover. The optimization objective focuses on the base policy:
\begin{equation}
    \max \mathcal{J}_{\mathrm{GRPO}}(\theta), \quad y \sim \pi_{\phi} 
\end{equation}

By alternating between these two stages, the exploration module continuously adapts to the evolving base policy, while the base policy gradually absorbs high-quality exploratory behaviors under a more stabilized training signal.



\subsection{Discussion}
\label{sec:method:discussion}

The premise of \method is that the rollout policy and the inference policy should be treated as two distinct objects with different design objectives: the inference policy is optimized for accurate, consistent, deployed responses, whereas the rollout policy is optimized for the diverse, exploratory, occasionally erroneous trajectories that yield informative gradients. Insisting on a single parameter set for both is the default in RL for LLM reasoning, but we view it as a default rather than a justified choice.

Approaches addressing the exploration-exploitation dilemma in Section~\ref{sec:ee}, including modified objectives, clipping strategies, and auxiliary sampling, still operate within a single policy whose parameters jointly govern training rollouts and inference responses, so the gradients that encourage exploration and those driving reward maximization remain entangled. \method instead confines the exploration signal to a separate Rollout-Head, so that gradients shaping the training distribution do not directly perturb the parameters producing deployed responses. We further discuss the relationship between \method and parameter-efficient fine-tuning (PEFT) in Appendix~\ref{sec:peft}.

\section{Experiments}

In this section, we conduct extensive experiments to evaluate the effectiveness of \method in enhancing reasoning capabilities and training stability. We begin by comparing \method against competitive baselines including GRPO and recent entropy-based exploration methods across two representative domains: mathematical reasoning and code generation. Subsequently, we provide a series of diagnostic analyses to examine the mechanistic drivers of \method. Finally, we discuss the computational efficiency and parameter overhead of the proposed framework to demonstrate its practical scalability.

\subsection{Experimental Setup}
We evaluate the performance of \method across diverse reasoning tasks. This section details our base models, benchmark datasets, and evaluation protocol.

\paragraph{Base Models} We utilize Qwen2.5-3B~\cite{qwen2025qwen25technicalreport} and the reasoning-enhanced Qwen3-8B~\cite{yang2025qwen3technicalreport} as backbones to evaluate the scalability of our method across different scales.

\paragraph{Datasets} Our evaluation spans two representative domains: both Mathematical Reasoning and Code Generation.
In the tasks of mathematical reasoning, we use GSM8K~\cite{DBLP:journals/corr/abs-2110-14168} for training and in-distribution testing, and MATH-500~\cite{lightman2024verifystepbystep} as an out-of-distribution benchmark to assess generalization in more sophisticated logical deduction.
For programming tasks, we employ MBPP~\cite{austin2021mbpp} for training, while evaluating functional synthesis capabilities on HumanEval~\cite{chen2021humaneval} and the multi-level algorithmic challenges posed by APPS~\cite{DBLP:conf/nips/HendrycksBKMAGB21}.

\paragraph{Baselines} Beyond GRPO, we include two entropy-based exploration methods closely related to our work.
\textbf{EPO}~\cite{DBLP:journals/corr/abs-2509-22576} augments GRPO with a policy-level entropy regularization term to prevent premature convergence.
\textbf{ETPO}~\cite{wen2024entropyregularizedtokenlevelpolicyoptimization} instead applies entropy regularization at the token level, incorporating per-token entropy into advantage estimation for finer-grained exploration.
Both modify the optimization objective and thus serve as direct comparisons against our structural decoupling approach. 

\paragraph{Metrics} For all benchmarks, we primarily report the Pass@1 accuracy. During training, we adopt a rule-based parser to extract numerical answers from \texttt{<answer>} tags for mathematical tasks, while for coding tasks, generated programs are validated through a standard execution sandbox to verify functional correctness. 

\paragraph{Evaluation Protocol} All benchmarks are evaluated using the OpenCompass framework \cite{opencompass2023}. To ensure a fair comparison and eliminate potential performance discrepancies caused by prompt sensitivity, we further customized the evaluation configurations to maintain strict consistency between the inference prompts and the training templates. \footnote{The rule-based metric provided by OpenCompass for MATH-500 is not fully compatible with the output format generated during our training; we therefore evaluate this dataset with a tailored metric.}

\subsection{Main Results} 

\begin{table*}[htbp]
\centering
\caption{Benchmark results comparison. Our proposed \textbf{\method} framework achieves consistent improvements over GRPO and entropy-based exploration baselines across both mathematical and programming reasoning tasks.} 
\label{tab:model_performance_refined}
\small
\setlength{\tabcolsep}{8pt} 
\resizebox{0.98\linewidth}{!}{
\begin{tabular}{ll cccccc  c} 
\toprule
\multirow{2}{*}{\textbf{Backbone}} & \multirow{2}{*}{\textbf{Method}} & \multicolumn{2}{c}{\textbf{Math (Pass@1)}} & \multicolumn{3}{c}{\textbf{Code (Pass@1)}} & \multirow{2}{*}{\textbf{Avg.}} \\
\cmidrule(lr){3-4} \cmidrule(lr){5-7}
& & GSM8K & MATH500 & MBPP & HumanEval & APPS & \\
\midrule
\multirow{5}{*}{Qwen2.5-3B}
& Base & 76.12 & 36.00 & 54.20 & 64.63 & 9.02 & 47.99 \\
& + GRPO & 80.74 & 42.20 & 58.60 & \textbf{68.90} & 8.22 & 51.73 \\
& + EPO~\cite{DBLP:journals/corr/abs-2509-22576} & 81.27 & 41.20 & 55.20 & 66.46 & 9.04 & 50.63 \\
& + ETPO~\cite{wen2024entropyregularizedtokenlevelpolicyoptimization} & 81.80 & 42.40 & 54.80 & 67.68 & 8.62 & 51.06 \\
& \textbf{+ \method (Ours)} & \textbf{83.17} & \textbf{45.60} & \textbf{59.00} & \textbf{68.90} & \textbf{9.52} & \textbf{53.24} \\
\midrule
\multirow{5}{*}{Qwen3-8B}
& Base & 80.74 & 51.00 & 58.80 & 66.46 & 12.84 & 53.97 \\
& + GRPO & 88.55 & 54.40 & 66.60 & 86.59 & 15.38 & 62.30 \\
& + EPO~\cite{DBLP:journals/corr/abs-2509-22576} & \textbf{89.92} & 56.80 & 63.80 & 85.98 & 15.50 & 62.40 \\
& + ETPO~\cite{wen2024entropyregularizedtokenlevelpolicyoptimization} & 89.46 & 55.40 & 66.60 & 87.20 & 15.14 & 62.76 \\
& \textbf{+ \method (Ours)} & 88.48 & \textbf{57.20} & \textbf{67.20} & \textbf{87.80} & \textbf{15.56} & \textbf{63.25} \\
\bottomrule
\end{tabular}}
\end{table*}

The comparative results presented in Table~\ref{tab:model_performance_refined} show that our proposed \textbf{\method} framework consistently outperforms the vanilla GRPO baseline across both the 3B and 8B model scales.

\paragraph{Superiority in Complex Reasoning}A key observation is that the performance gains of \method are more pronounced on challenging, out-of-distribution benchmarks. For instance, on the \texttt{MATH-500} dataset, the Qwen2.5-3B model with our method achieves an absolute improvement of \textbf{3.4\%} over the standard GRPO (45.6\% vs. 42.2\%), and the Qwen3-8B model sees a similar gain of \textbf{2.8\%}. Similarly, in the \texttt{APPS} coding benchmark, which requires sophisticated algorithmic synthesis, \method achieves the highest Pass@1 scores across both backbones. This suggests that the decoupled Rollout-Head successfully encourages diverse, advanced reasoning paths critical for complex multi-step logical deduction. \method also outperforms EPO and ETPO on both backbones, demonstrating the superiority of architectural decoupling over objective-level entropy regularization.

\paragraph{Decoupling as a Stability Buffer}While standard GRPO shows a performance decline in some coding tasks (e.g., Qwen2.5-3B on \texttt{APPS} dropping from 9.02 to 8.22), our \method maintains or even enhances the base model's performance. This supports our hypothesis regarding the representational conflict. In vanilla GRPO, the destructive gradient noise from high-variance exploration may destabilize the learned patterns in the primary head. In contrast, the Rollout-Head acts as a structural buffering mechanism. This architectural isolation ensures that the core policy can exploit stable reasoning paradigms while the rollout branch probes the fringes of the solution space.

\paragraph{Scalability and Generalization} Even when the base policy is already strong, the Rollout-Head continues to provide incremental gains. The fact that the base policy ($\pi_{\theta}$) reaches higher accuracy after being trained with trajectories sampled from the exploration policy ($\pi_{\phi}$) confirms the effectiveness of our \textit{Two-Stage Iterative Training}. The exploration head successfully discovers rare but correct reasoning paths that effectively expand the cognitive boundaries of the primary backbone.

\subsection{Stability and Diversity Analysis}

In this section, we analyze the stability and diversity of \method from three perspectives:
(1) reward dynamics and stability,
(2) rollout diversity,
(3) length bias and verbosity.

\subsubsection{Reward Dynamics and Stability} 
\begin{figure}[!t]
    \centering
        \centering
        \includegraphics[width=0.95\linewidth]{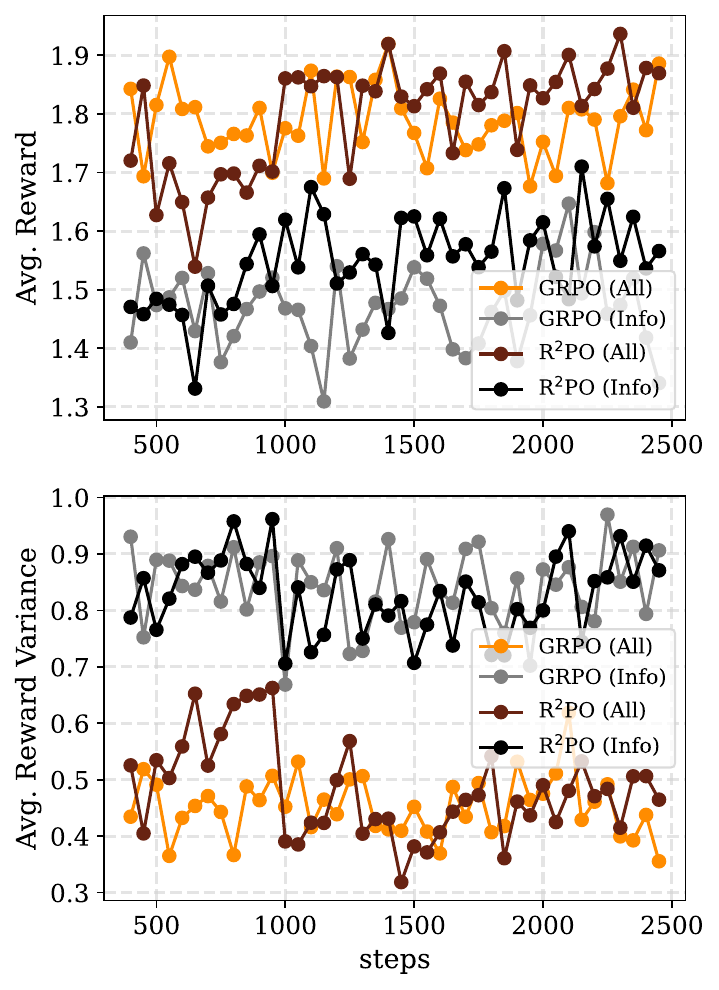}
    \caption{The mean and variance of rewards during the training process on \texttt{GSM8K}. The upper row shows the mean reward, while the lower row displays the variance of the mean reward. “Info” refers to Informative Rollouts, which contain varying rewards and thus supply non-zero advantage signals for policy updates.}
    \label{fig:training_reward}
\end{figure}

We monitor the mean and variance of rewards during the training process on \texttt{GSM8K}. Specifically, we distinguish between the global reward mean of all generated trajectories and the subset of Informative Rollouts defined as groups where reward variance exists, thereby providing non-zero advantage signals for policy updates. As illustrated in Figure \ref{fig:training_reward}, our model consistently achieves a higher mean reward with a more pronounced upward trajectory compared to the baseline GRPO, while keeping the reward variance essentially comparable. This suggests that the decoupled architecture facilitates a more thorough search of the solution space before converging to a stable, high-performing policy.

\subsubsection{Rollout Diversity}
To examine whether decoupled training actually shapes a more diverse trajectory distribution, we cluster rollouts on \texttt{GSM8K} in batches of 64 with HDBSCAN~\cite{mcinnes2017hdbscan} over sentence embeddings, and average per-batch statistics over the full set of rollouts using \texttt{BAAI/bge-large-en-v1.5} ~\cite{xiao2024cpack}. We report three quantities: \textbf{clusters}, the number of valid HDBSCAN clusters per batch (excluding the noise label); \textbf{noise}, the number of samples per batch that fit no dense mode; and \textbf{entropy}, the Shannon entropy $H = -\sum_{c} \tfrac{|c|}{N} \log \tfrac{|c|}{N}$ of the cluster-size distribution (with $N$ the total batch size including noise). We compare the Qwen2.5-3B base model, GRPO at step 2500, and \method at step 2500, drawing 64 rollouts per prompt over the entire test set.

\begin{table}[htbp]
\centering
\caption{Rollout diversity on \texttt{GSM8K} (Qwen2.5-3B). \method produces more semantic clusters, fewer unstructured noise points, and higher cluster entropy than GRPO at the same training step.}
\label{tab:cluster}
\setlength{\tabcolsep}{4pt}
\begin{tabular}{lccc}
\toprule
\textbf{Model} & \textbf{clusters} & \textbf{noise} & \textbf{entropy} \\
\midrule
BASE  & 2.459 & 34.659 & 0.5429 \\
GRPO  & 2.522 & 26.346 & 0.5782 \\
\method & \textbf{2.707} & \textbf{24.942} & \textbf{0.6136} \\
\bottomrule
\end{tabular}
\end{table}

\method simultaneously increases the number of semantic modes, reduces unstructured noise, and raises cluster entropy, while GRPO improves over the base only marginally. This is direct evidence for our thesis: routing exploration through a separate Rollout-Head expands the \emph{optimization-optimal} trajectory distribution into more, better-populated modes rather than collapsing onto a narrow set of dominant responses.

\subsubsection{Length Bias and Verbosity} 
Standard GRPO often suffers from a "verbosity bias" where incorrect answers tend to increase in length to minimize token-level penalties \cite{DBLP:journals/corr/abs-2501-12948}. In contrast, our model produces more concise responses. We compare the average length of correct and incorrect answers against a baseline GRPO model from the early training phase (100 steps) to serve as a reference point, as its outputs are sufficiently standardized to avoid evaluation bias under our metrics. 
Our results indicate that by decoupling exploration, the model avoids the trap of using excessive tokens as a buffer for negative gradients, leading to more efficient reasoning paths.

\begin{table}[htbp]
\centering
\caption{Average Response Length (tokens) Comparison. Our model suppresses the "length explosion" observed in late-stage GRPO, particularly for incorrect trajectories. Measured on the first 100 samples of the GSM8K test set.}
\label{tab:length_analysis}
\resizebox{0.8\linewidth}{!}{
\begin{tabular}{lccc}
\toprule
\textbf{Response Type} & \textbf{Base} & \textbf{GRPO} & \textbf{\method} \\
\midrule
Correct Answers   & 211.61  & 211.95 & 204.12 \\
Incorrect Answers & 281.22 & 295.28 & 281.59 \\
\bottomrule
\end{tabular}}
\end{table}

Beyond the three analyses above, we further verify in Appendix~\ref{sec:passk} that early-stage Pass@8 is not degraded by the additional exploration mechanism, and in Appendix~\ref{sec:misspec} that \method is immune to a reward-misspecification perturbation that GRPO succumbs to (0\% vs 0.875\% format-error rate).

\subsection{Ablation Study}
\label{sec:analysis}


To identify what truly drives the gains of \textbf{\method}, we isolate its structural and reward-level components via two ablations: (i) the role of the GIF reward signal, and (ii) the necessity of the Rollout-Head.

\paragraph{Reward Signal}
Instead of the Group Inverse-Frequency (GIF) Reward, we train the Rollout-Head using the main reward, i.e., the same accuracy and formatting rewards as the primary policy used. This ablation evaluates whether the improvements come mainly from the \emph{architectural decoupling} itself rather than the \emph{Group Inverse-Frequency Reward} signal.
As shown in Table~\ref{tab:ablation}, the Correctness and Formatting Reward configuration achieved the highest overall performance, slightly exceeding the Group Inverse-Frequency Reward variant. This suggests that the structural separation of exploration is the primary driver of success, providing a stable sandbox for the model to refine reasoning paths. This is an encouraging finding, as it implies that \method can be successfully deployed without complex reward engineering. 

\paragraph{Residual Branch}
To evaluate if the gains simply result from the staggered frozen training schedule, we implemented a version without the Rollout-Head, where we periodically froze the LM Head while updating the backbone, and vice-versa. 
As also shown in Table~\ref{tab:ablation}, the optimization without the Rollout-Head outperformed vanilla GRPO but lagged behind the full \method, confirming that the additional parameter capacity of the residual branch is essential for effective decoupling.

\begin{table}[htbp] 
\centering 
\caption{Ablation Study on \texttt{GSM8K} using Qwen2.5-3B} 
\label{tab:ablation} 
\resizebox{0.8\linewidth}{!}{
\begin{tabular}{lcc} 
\toprule 
\textbf{Configuration} & \textbf{GSM8K (Math)} \\ 
\midrule Vanilla GRPO & 80.74  \\ 
\method  (w/o Rollout-Head) & 82.18 \\
\method (w/ GIF Reward) & 83.17  \\
\method (w/ Main Reward) & \textbf{83.55} \\
\bottomrule 
\end{tabular}}
\end{table}

\subsection{Efficiency and Parameter Discussion}
To evaluate the computational efficiency of \method, we analyze the parameter overhead and its impact on both training and inference phases. As shown in Table~\ref{tab:params}, \method introduces a minor training parameter overhead (7.8\%–10.2\%), which is detached during inference to ensure zero additional latency.
\begin{table}[h]
\centering
\caption{Parameter overhead across model scales.}
\label{tab:params}
\setlength{\tabcolsep}{4pt}
\resizebox{0.9\linewidth}{!}{
\begin{tabular}{lcccc}
\toprule
\textbf{Model} & \textbf{Base} & \textbf{Full} & \textbf{Added ($\Delta$)} & \textbf{Growth (\%)} \\
\midrule
Qwen2.5-3B & 3.09 B & 3.40 B & 315.36 M & $\sim$10.2\% \\
Qwen3-8B & 8.19 B & 8.83 B & 639.11 M & $\sim$7.8\% \\ 
\bottomrule
\end{tabular}}
\end{table}

Beyond parameter count, we further examine the peak GPU memory usage during training. Compared to vanilla GRPO, which requires 58,519 MB of GPU memory, \method exhibits a markedly different memory profile across its two stages. In Stage~1, where only the Rollout-Head is optimized and the backbone is frozen, the peak memory usage drops significantly to 26,343 MB, representing a reduction of over 55\%. In Stage~2, where the backbone and LM Head are updated while the Rollout-Head is frozen, the memory usage (60,563 MB) is comparable to that of vanilla GRPO. The proposed framework therefore incurs no additional inference cost and only marginal training overhead, making it suitable for large-scale deployment.

\section{Conclusion}
We argue that tying the rollout policy to the inference policy is a misleading inductive bias in RL for LLM reasoning: the \emph{optimization-optimal} and \emph{inference-optimal} distributions serve different roles and should not share one parameter set. \method realizes this view minimally via a discardable residual Rollout-Head atop a frozen inference policy. Across mathematical reasoning and code generation, this decoupling increases rollout diversity, controls length bias, and improves accuracy and stability over GRPO, suggesting future methods treat the two policies as distinct design objects.

\section*{Limitation}
\method is evaluated only on models of 3B and 8B parameters. Its behavior on substantially larger models (e.g., 70B+) or smaller-scale models remains to be explored. For large-scale models, it is unclear whether the Rollout-Head can continue to yield meaningful gains with a relatively small number of additional parameters. Conversely, for smaller models, the parameter increase introduced by the Rollout-Head, although minimized, may still constitute a non-negligible overhead given their limited capacity and vocabulary size.

Moreover, our experiments focus exclusively on single-turn reasoning. Extending this decoupled exploration framework to multi-turn dialogue settings or long-horizon agent tasks may introduce additional challenges, which we leave for future investigation.

Finally, the GIF reward used in our experiments is tailored to discrete reward settings, and its extension to dense or continuous reward landscapes remains open. We note that the decoupled architecture itself is reward-agnostic: our ablations show that even when the Rollout-Head is trained with the main task reward alone, \method retains its advantage over GRPO. Designing exploration rewards better matched to continuous-reward regimes is an interesting direction for future work. 

\section*{Ethics Statement} While \method enhances the reasoning and problem-solving capabilities of LLMs, we recognize potential risks associated with its deployment. Improved reasoning could theoretically be misused for generating sophisticated malicious code or automated misinformation. To mitigate such risks, our experiments rely solely on publicly available datasets, and the training follows standard RLHF safety alignment practices.  Moreover, the structural decoupling introduced in \method improves optimization transparency and model interpretability, which can help in auditing model behavior and reducing unintended outputs. We emphasize that the deployment of \method should consider these risks and be accompanied by appropriate monitoring and access control.

\section*{Use of AI Assistants}
AI assistants (specifically ChatGPT and Claude) were used in this research primarily for language polishing, improving grammatical structures, and refining the clarity of the manuscript. All original methodology, experimental design, and data analysis were conducted solely by the human authors.


\bibliography{custom}

\appendix
\section{Appendices}

\subsection{The Details of Datasets}
We evaluate the performance of \method across several high-quality reasoning and coding benchmarks. The statistics of these datasets are summarized in Table~\ref{tab:dataset_stats}.

\textbf{GSM8K \cite{DBLP:journals/corr/abs-2110-14168}} The Grade School Math 8K dataset consists of 8.5K high-quality grade school math word problems. We use the training set for reinforcement learning and the test set for in-distribution evaluation. This benchmark tests basic multi-step arithmetic reasoning.

\textbf{MATH-500 \cite{lightman2024verifystepbystep}} A subset of the original MATH dataset, containing 500 challenging problems across several subjects like algebra and geometry. We use this as an out-of-distribution (OOD) test set to evaluate the generalization of models trained on GSM8K.

\textbf{MBPP \cite{austin2021mbpp}} The Mostly Basic Python Problems dataset contains around 1,000 entry-level Python programming problems. It is used to fine-tune the model's basic code generation capabilities through verifiable rewards.

\textbf{HumanEval \cite{chen2021humaneval}} A standard benchmark consisting of 164 handwritten Python programming problems. It evaluates the model's ability to solve functional programming tasks based on docstrings.

\textbf{APPS \cite{DBLP:conf/nips/HendrycksBKMAGB21}} The Automated Programming Progress Standard contains 10,000 coding challenges ranging from introductory to competition level. It assesses sophisticated algorithmic synthesis and is a key metric for reasoning-intensive coding tasks.

\begin{table}[htbp] 
\centering 
\small 
\caption{Statistics of the datasets used for \method training and evaluation.} 
\label{tab:dataset_stats} 
\begin{tabular}{lccc} 
\toprule 
\textbf{Dataset} & \textbf{Domain} & \textbf{Train Size} & \textbf{Test Size} \\ 
\midrule 
GSM8K & Math & 7,473 & 1,319 \\
MATH-500 & Math & - & 500 \\
MBPP & Code & 374 & 500 \\ 
HumanEval & Code & - & 164 \\ 
APPS & Code & - & 5,000 \\ 
\bottomrule 
\end{tabular} 
\end{table}

\subsection{The Details of Baselines}
We select two models from the Qwen series to evaluate \method across different parameter scales and capability levels:

\textbf{Qwen2.5-3B \cite{qwen2025qwen25technicalreport}} A highly efficient language model with 3.09 billion parameters. It serves as our primary testbed for analyzing training dynamics and performing ablation studies due to its balanced performance and computational efficiency.

\textbf{Qwen3-8B \cite{yang2025qwen3technicalreport}} The latest reasoning-enhanced backbone with 8.19 billion parameters. This model features a deeper understanding of complex logical structures and improved long-context reasoning capabilities. We use this model to verify the scalability of \method and its effectiveness when applied to stronger base policies.

\subsection{Training Hyperparameters}
All experiments are conducted on a cluster of NVIDIA A800-SXM4-80GB GPUs. For both the Qwen2.5-3B and Qwen3-8B backbones, we employ the AdamW optimizer with a constant learning rate of $1 \times 10^{-6}$ and a cosine decay schedule. To ensure a fair comparison, \texttt{GRPO}\xspace and \method share identical RL configurations: a group size ($G$) of 8, a KL-divergence coefficient ($\beta$) of 0.04, and a PPO clip range ($\epsilon$) of 0.2. The maximum sequence length is set to 512 for 3B models and 768 for 8B models. During the first 2000 training steps, we use 150 steps for Stage~1 and 200 steps for Stage~2 per cycle; afterwards, the cycle is lengthened to 200 and 300 steps respectively. This schedule is fixed across all tasks without task-specific tuning. 


\subsection{Comparison with Parameter-Efficient Fine-Tuning}
\label{sec:peft}

The Rollout-Head in \method is architecturally close to lightweight modules used in Parameter-Efficient Fine-Tuning (PEFT) approaches such as Adapters~\cite{DBLP:conf/icml/HoulsbyGJMLGAG19} and LoRA~\cite{DBLP:conf/iclr/HuSWALWWC22}, in that all three attach a small trainable branch to a frozen or partially frozen backbone. Its functional role, however, is fundamentally different along three axes.

\paragraph{Optimization objective.} PEFT methods aim to reduce the number of trainable parameters while preserving the original optimization objective and trajectory distribution. The adaptation modules learn to approximate a full fine-tune of the backbone, and the trajectories used to update them are sampled from the same policy that the adapted model will deploy. In contrast, the Rollout-Head in \method is introduced precisely to reshape the sampling distribution used for policy optimization, so that the effective optimization objective is altered even though the loss function remains the standard group-relative policy optimization objective. Whereas PEFT moves the parameter footprint, \method moves the trajectory distribution.

\paragraph{Role at training versus inference.} In PEFT, the adaptation module is trained jointly with (or on top of) the backbone and remains active at inference; it is part of the deployed model. In \method, the Rollout-Head is used only during training as the behavior policy that generates exploratory rollouts, and is discarded at inference, so the deployed model is architecturally identical to a standard LLM. The Rollout-Head therefore changes the optimization geometry but leaves no inference-time footprint, which is the opposite of what PEFT is designed for.

\paragraph{Exploration versus exploitation specialization.} The Rollout-Head and the backbone in \method are deliberately specialized: the Rollout-Head, trained with a rarity-driven exploration signal, expands the trajectory distribution toward informative-gradient regions, while the backbone, trained on these trajectories with the task reward, consolidates stable high-reward reasoning patterns. PEFT modules are not designed to support such a role separation; they share gradients with the backbone under a single objective. The two paradigms are therefore orthogonal rather than competing: \method can in principle be combined with PEFT (for example, by realizing the Rollout-Head itself as an adapter), but they address different axes, namely parameter efficiency versus trajectory-distribution control.

\subsection{Early-Stage Pass@k Comparison}
\label{sec:passk}

Replay-based exploration methods are known to occasionally degrade early-stage Pass@k performance, because reused off-policy trajectories may distort the current policy's sampling distribution. Since \method generates trajectories with an on-policy behavioral sampler (the Rollout-Head) rather than reusing past data, it does not suffer from this issue. To verify this empirically, we report Pass@8 at two early training checkpoints on \texttt{GSM8K}.

\begin{table}[htbp]
\centering
\caption{Early-stage Pass@8 on \texttt{GSM8K} (Qwen2.5-3B). \method maintains comparable or slightly higher Pass@8 throughout early training.}
\label{tab:passk_early}
\setlength{\tabcolsep}{8pt}
\begin{tabular}{lcc}
\toprule
\textbf{Method} & \textbf{200 steps} & \textbf{400 steps} \\
\midrule
GRPO & 0.9204 & 0.9287 \\
\method & 0.9227 & 0.9295 \\
\bottomrule
\end{tabular}
\end{table}

The results confirm that the additional exploration mechanism does not harm early-stage sampling diversity, which is consistent with the on-policy nature of \method.

\subsection{Robustness to Reward Misspecification}
\label{sec:misspec}

A critical challenge in rule-based reinforcement learning is the gap between the training reward function and the actual evaluation criteria. We encountered a case of Reward Misspecification: during training, the format reward was loosely defined (requiring only the presence of \texttt{<think>} and \texttt{<answer>} tags), whereas the evaluation script strictly enforced a single occurrence of these tags.

In the baseline GRPO, we observed a sudden spike in reward variance on the test set between steps 2500 and 3000, exceeding even the initial exploration phase. As shown in \ref{sec:case}, diagnostic analysis revealed that the model learned to output redundant \texttt{<think>} blocks, which is a behavior that was not penalized during training but caused catastrophic failures during evaluation. According to the error evolution trends in Table \ref{tab:error_evolution}, this noise or sub-optimal behavior is an emergent artifact of prolonged training. We utilize the GRPO model at step 100 as a baseline, as it represents the stage where the model has fundamentally mastered the necessary formatting requirements to obtain rewards.

\begin{table}[htbp]
\centering
\caption{Evolution of Format Error Rate in Baseline GRPO. The noise (redundant \texttt{<think>} tags) emerges and accelerates during the late stages of training due to reward misspecification. Measured on the first 100 samples of the GSM8K test set.}
\label{tab:error_evolution}
\setlength{\tabcolsep}{4pt}
\begin{tabular}{lcccc}
\toprule
\textbf{Training Steps} & 100 & 2000 & 2500 & 3000 \\
\midrule
\textbf{Error Rate (\%)} & 0.125 & 2.750 & 6.875 & 46.625 \\
\bottomrule
\end{tabular}
\end{table}

To test whether our model's resilience to this issue was coincidental, we conducted a \textbf{Perturbation Experiment}. After 2000 steps, we manually injected redundant tags into successful trajectories for 10 steps to simulate a "noise trap." We then measured the probability of the model adopting this erroneous behavior in the subsequent 100 steps.

\begin{table}[htbp]
\centering
\caption{Results of the Perturbation Experiment. Measured on the first 100 samples of the GSM8K test set.}
\label{tab:perturbation_results}
\resizebox{0.7\linewidth}{!}{
\begin{tabular}{lcc}
\toprule
\textbf{Steps} & \textbf{GRPO} & \textbf{\method} \\
\midrule
2000 (Start) & 0.125\% & \textbf{0.000\%} \\
2050         & 0.375\% & \textbf{0.000\%} \\
2100         & 0.875\% & \textbf{0.000\%} \\
\bottomrule
\end{tabular}}
\end{table}

The baseline GRPO quickly succumbed to the perturbation, with the error rate rising to 0.875\%, whereas our model maintained a 0\% error rate. This clearly demonstrates that our decoupled head architecture provides significantly higher \textbf{perturbation immunity}, likely because the primary policy is effectively shielded from the high-variance exploration noise by the Rollout-Head.

\subsection{Case Study: Redundant Reasoning Patterns in GRPO}
\label{sec:case}

As shown in the example, the standard GRPO model frequently produces redundant \texttt{<think>} tags and repetitive logic cycles. This behavior indicates that without the structural decoupling provided by \method, the single-head policy are more susceptible to undesirable reward noise.

\begin{table*}[b]
    \centering
    \footnotesize
    \caption{Case Study on Formatting and Verbosity. The baseline GRPO model produces an over-optimized response with repetitive \think{} tags, which often leads to parsing failures in strict evaluation environments. \method effectively suppresses this formatting noise through its decoupled architecture, yielding a significantly shorter response (one \think{} block) without compromising mathematical accuracy.}\label{tab:case}
    \begin{tabular}{p{13.5cm}}
\specialrule{.15em}{.5em}{.5em}
\textbf{Case 1} \\
\hline
\textbf{Question}: Toulouse has twice as many sheep as Charleston. Charleston has 4 times as many sheep as Seattle. How many sheep do Toulouse, Charleston, and Seattle have together if Seattle has 20 sheep? \\
\hline
\textbf{Ground Truth}: 260 \\
\hline
\texttt{GRPO}\xspace: \\
\think{First, let's find out how many sheep Charleston has. Since Toulouse has twice as many sheep as Charleston, and Charleston has 4 times as many sheep as Seattle, we can start by finding out how many sheep Charleston has: Charleston sheep = 20 (Seattle's sheep) * 4 = 80 sheep}  \\ 
\think{Now, to find the number of sheep in Toulouse, we multiply Charleston's number of sheep by 2: Toulouse sheep = 80 (Charleston's sheep) * 2 = 160 sheep.} \\ 
\think{Now, let's sum up the total number of sheep in Toulouse, Charleston, and Seattle: Total sheep = Toulouse sheep + Charleston sheep + Seattle sheep = 160 + 80 + 20 = 260 sheep.} \\ 
\answer{The total number of sheep in Toulouse, Charleston, and Seattle is 260.} \\
\hline
\textbf{\method}: \\
\think{Let's start by figuring out how many sheep each city has based on the information provided. If Seattle has 20 sheep, and Charleston has 4 times as many sheep as Seattle, then Charleston has 4 * 20 = 80 sheep. Toulouse has twice as many sheep as Charleston, so Toulouse has 2 * 80 = 160 sheep. To find the total number of sheep in Toulouse, Charleston, and Seattle together, we add the number of sheep in each city: 20 + 80 + 160 = 260 sheep} \\
\answer{ 260 sheep } \\
\hline
\specialrule{.15em}{.5em}{.5em}
\textbf{Case 2} \\
\hline
\textbf{Question}: Kylar went to the store to buy glasses for his new apartment. One glass costs \$5, but every second glass costs only 60\% of the price. Kylar wants to buy 16 glasses. How much does he need to pay for them? \\
\hline
\textbf{Ground Truth}: 64 \\
\hline
\texttt{GRPO}\xspace: \\
\think{First, let's calculate the cost of one glass, which is \$5.} \\
\think{For every second glass, Kylar will pay 60\% of \$5, which is 0.6 * \$5 = \$3.}\\
\think{Kyle wants to buy 16 glasses. In this case, he will buy 8 glasses for the first price and 8 glasses for the second price (60\% of the price).} \\
\think{So, the total cost for the first 8 glasses is 8 * \$5 = \$40.} \\
\think{The total cost for the second 8 glasses is 8 * \$3 = \$24.} \\
\think{Therefore, the total cost for Kylar's glasses is \$40 + \$24 = \$64.} \\
\answer{\$64} \\
\hline
\textbf{\method}: \\
\think{First, Kylar needs to buy 16 glasses. The price for the first glass is \$5. For every second glass, he pays 60\% of \$5, which is 0.60 * \$5 = \$3. So, for every pair of glasses (one original and one discounted), he pays \$5 + \$3 = \$8. Since he wants to buy 16 glasses, half of them (8 pairs) are original, and the other half (8 pairs) are discounted. Therefore, the total cost is 8 * \$5 + 8 * \$3 = \$40 + \$24 = \$64.} \\
\answer{ \$64 } \\
\hline
    \end{tabular}
\end{table*}

\end{document}